\documentclass[10pt,journal]{IEEEtran}
\usepackage{amsmath,amsfonts}
\usepackage{algorithmic}
\usepackage{array}
\usepackage[caption=false,font=normalsize,labelfont=sf,textfont=sf]{subfig}
\usepackage{textcomp}
\usepackage{stfloats}
\usepackage{url}
\usepackage{bm}
\usepackage{cuted}
\usepackage{longtable}
\usepackage{verbatim}
\usepackage{graphicx}
\usepackage{amsmath}
\usepackage{lineno,hyperref}
\usepackage{multirow}
\usepackage{makecell}
\usepackage{pdflscape}
\usepackage{rotating}
\usepackage{booktabs} 
\usepackage{tabularx} 
\usepackage{caption}  

\usepackage{stfloats}
\def\BibTeX{{\rm B\kern-.05em{\sc i\kern-.025em b}\kern-.08em
    T\kern-.1667em\lower.7ex\hbox{E}\kern-.125emX}}
\usepackage{balance}
\begin{document}
\title{Transformer-based EEG Decoding: A Survey}
\author{Haodong Zhang, Hongqi Li*~\IEEEmembership{Member,~IEEE}
\thanks{Manuscript received July 02, 2025. This work was supported in part by the Natural Science Basic Research Program of Shaanxi Province under Grant 2024JC-YBQN-0659. (Corresponding author: Hongqi Li.) 

H. Zhang and H. Li are with the School of Software, Northwestern Polytechnical University, Xi’an 710072, China, (e-mail: zhang\_haodong@mail. nwpu.edu.cn, lihongqi@nwpu.edu.cn).}}

\markboth{Journal of \LaTeX\ Class Files,~Vol.~18, No.~9, September~2020}%
{How to Use the IEEEtran \LaTeX \ Templates}

\maketitle
\begin{abstract}
Electroencephalography (EEG) is one of the most common signals used to capture the electrical activity of the brain, and the decoding of EEG, to acquire the user intents, has been at the forefront of brain-computer/machine interfaces (BCIs/BMIs) research. Compared to traditional EEG analysis methods with machine learning, the advent of deep learning approaches have gradually revolutionized the field by providing an end-to-end long-cascaded architecture, which can learn more discriminative features automatically. Among these, Transformer is renowned for its strong handling capability of sequential data by the attention mechanism, and the application of Transformers in various EEG processing tasks is increasingly prevalent. This article delves into a relevant survey, summarizing the latest application of Transformer models in EEG decoding since it appeared. The evolution of the model architecture is followed to sort and organize the related advances, in which we first elucidate the fundamentals of the Transformer that benefits EEG decoding and its direct application. Then, the common hybrid architectures by integrating basic Transformer with other deep learning techniques (convolutional/recurrent/graph/spiking neural netwo-rks, generative adversarial networks, diffusion models, etc.) is overviewed in detail. The research advances of applying the modified intrinsic structures of customized Transformer have also been introduced. Finally, the current challenges and future development prospects in this rapidly evolving field are discussed. This paper aims to help readers gain a clear understanding of the current state of Transformer applications in EEG decoding and to provide valuable insights for future research endeavors.
\end{abstract}


\captionsetup{
  labelsep=space,     
}
\section{Introduction}
\label{chap:1}
\IEEEPARstart{A}{S} a cutting-edge technology to directly bridge the human mind with external devices, brain-computer/machine interfaces (BCIs/BMIs) have garnered more and more attention in both rehabilitation engineering and non-medical domains \cite{1,2}. The non-invasive electroencephalography (EEG) records the activity of cortical neurons through electrodes placed on the scalp, providing millisecond-level temporal resolution that allows for tracking rapid changes in brain activity, and has been widely used in BCIs \cite{3}. Accordingly, the efficient EEG decoding through the analysis to capture the underlying user intents is of great significance, which generally follows a set of steps with signal acquisition, preprocessing, feature extraction, and classification.

Traditionally, machine learning methods are used in the above procedures, where researchers usually start by extracting features manually of three main dimensions, i.e., frequency, temporal, and spatial domains from raw EEG signals, which are then fed into specific classifiers for decoding and recognition. Specifically, the features of the frequency, time-frequency, and spatial-frequency have been designed first based on the specific expertise and extracted with the common spatial pattern (CSP), Fast Fourier Transform (FFT), and Wavelet Transform, etc. Then, supervised classification algorithms such as the support vector machine, random forest, and linear discriminant analysis, or unsupervised learning methods such as K-nearest neighbor analysis (KNN) are employed. However, manual feature design and extraction rely heavily on expert experience and is a time-consuming and complex process. 

Conversely, deep learning (DL) methods have recently achieved great success in the field of EEG decoding due to their powerful automatic feature extraction and representation learning capabilities. Briefly, DL is inspired by a hierarchical structure based on the visual cortex of the human brain, which mainly represents a long cascade of information extraction architecture with multiple processing layers to reflect its depth. Since it allows for increasingly abstract but more representative features through the constant transfer of information from the lowest layer to the highest, DL techniques such as convolu¬tional neural networks (CNNs), recurrent neural networks (RNNs), long short-term memory (LSTM), transfer learning, have already been successfully applied for the EEG decoding. Nevertheless, many of above DL models struggle to parallelly model the complex temporal patterns and high-dimensional feature spaces of non-stationary EEG data.

As another classical type of DL algorithm, Transformer was introduced by Google in 2017 to revolutionize sequence processing tasks \cite{4}. The core innovation of Transformer is the multi-head attention mechanism (MHA), which calculates the relationships between elements in the input sequence and assigns different importance weights to each element. To this end, it enables the model to capture complex temporal and spatial patterns, particularly those long-range dependencies entailed in EEG signals. The positional encoding embedded in the model ensures accurate temporal positioning, which is essential for identifying intricate relationships across different segments of EEG sequences. Moreover, multiple attention heads also allow the Transformer to process information in parallel across different subspaces, enhancing both efficiency and expressive power. In a nutshell, the Transformer architec-ture bears the benefits of parallelization, scalability, and flexibility, which has shown a promising future in various domains \cite{5}. Since 2019, such a model has gradually appeared in EEG speech recognition \cite{6}, epilepsy detection \cite{7}, and classification tasks \cite{8}, and the resulting significant advances have led to a new wave of EEG research.

interfaces (BCIS/BMIS) have garnered much attention over the past decades due to their outstanding ability to convert the users’ brain activity into machine-readable intentions or commands \cite{1}. Among various BCI modalities, noninvasive electroencephalograph (EEG) has the advantages of adequate temporal resolution, non-surgical electrode placements, and low cost, thus leading to its widest application in the fields of rehabilitation engineering \cite{2,3}, cognitive science \cite{4}, neuroscience, and psychology \cite{5}. 

However, although nearly two hundred papers have been published to date, few comprehensive review articles were found to clarify the trends of the Transformers application in the EEG field. Chen et al. \cite{9} reported the use of fundamental Transformer model in the field of brain science, emphasizing on its broad variety of applications of the brain disease diagnosis, brain age prediction, brain anomaly detection, etc. More focally, in \cite{10}, Abibullaev et al. concluded the model’s foundational principles, technical advantages, and potential challenges applied to BCIs. Further, the importance of data limitations and augmentation in Transformer-based BCI systems has been highlighted in \cite{11}. However, to the best of our knowledge, there is no detailed overview of the specific evolution and variants of the relevant models to shed light on current trends in technological progress under the explosive growth of Transformers in EEG decoding. The urgent need for a thorough comprehensive summary of the recent advances gives rise to our current work.

\vspace{-0.4cm}
\subsection{Contribution and Overview}
We searched the literature from databases of IEEE Xplore, Science Direct, Web of Science, Google Scholar, and arXiv. The cut-off date of the screening was Jun. 29th, 2025, and the main keywords were “Electroencephalogram/EEG”, and “Transformer/vision transformer/ViT/self-attention/multi-head attention”, and “decoding/classification/detection/recognition”. After the search, the following screening criteria were applied:
\begin{itemize}
\item Studies focused on non-EEG signals (e.g., EOG, EMG) were excluded, except for involved multimodal methods.
\item Papers only mentioned Transformers in the title/abstract but were unrelated to the main text or lacked innovative methods were eliminated.
\item Studies with unclear design, insufficient result analysis, or lacking comparative experiments were disqualified.
\item Priority was given to influential studies validated using diverse standard datasets.
\end{itemize}

As a result, about 165 papers related to the application of Transformers in EEG decoding since 2020 were screened and compiled, to provide a comprehensive summary and in-depth analysis of recent advancements. The main contributions of this review are: 1) Provide readers with the most thorough overview to date on the applicability of Transformer model in the field of EEG decoding. The distillation carried out via this work facilitates clarifying the evolving lineage with relevant research and helps to excite the attention of potential researchers; 2) We classify the Transformer-based variants for EEG processing tasks in terms of direct Transformer, its combination with other DL approaches, and customization. Further, for the customized ones, we sort out for the first time the specific categories of enhanced intrinsic structures while offering clear research entry points; 3) Challenges and future directions of the Transformer within the EEG decoding are discussed in explicit detail, especially those applicable models related to specific decoding tasks and datasets, to inspire the further dialectical rethinking about the development of the field and flourish the community.
\vspace{-0.35cm}
\subsection{Paper Organization}
Overall, for a variety of EEG tasks (classification, generation, and denoising, etc.), the Transformer is exploited by means of the backbone models (used independently), hybrid models (combined with other DL-based networks), and customized Transformer (variants based on modified intrinsic structures). Hence, the remainder of paper is organized as follows. Section \ref{chap:2} provides an overview of Transformer basics and the direct application of backbone models to EEG decoding. Afterwards, along with the line of model evolution, current research are systematically reviewed, including the hybrid architectures of base Transformer with other DL models in Section \ref{chap:3}, and customized Transformer-based networks in Section \ref{chap:4}. Finally, Section \ref{chap:5} discusses current challenges and future directions, while Section VI concludes the current work.

\section{Transformer Basics and Direct Applications}
\label{chap:2}
The basic Transformer model optimizes the sequence-to-sequence tasks with its Encoder-Decoder structure and multi-head self-attention mechanism. These features have enabled the Transformer to effectively capture subtle patterns and dynamic changes of EEG, thus recognizing intricate patterns. 

\begin{figure}[htpb]
\centering
\includegraphics[width=8.2cm]{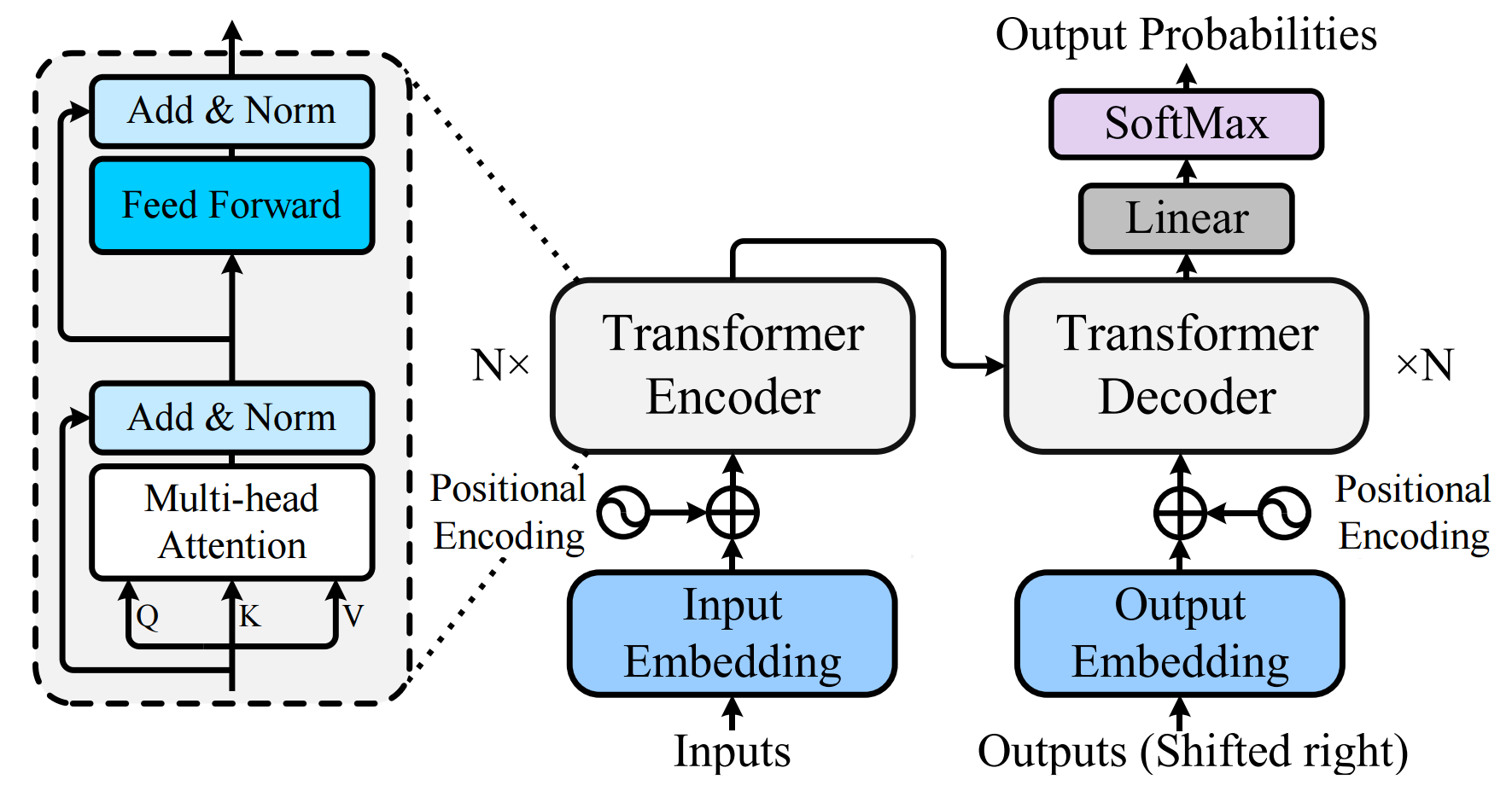}
\captionsetup{font=footnotesize}
\caption{General structure of Transformer backbone model.}
\label{fig 1}
\end{figure}
\vspace{-0.2cm}
\textbf{Fig.}\ref{fig 1} shows the general structure of Transformer backbone model, starting from Input Embedding, which provides a numerical representation of raw data by converting it into dense vectors. Then, Positional Encoding adds unique position-based information to preserve sequence order. Two core parts of the Transformer Encoder and Decoder consist of multiple layers to construct learning models, where the self-attention mechanism of Encoder processes the entire sequence context and Decoder employs masked attention to preserve causality in sequence generation tasks. The core contents of self-attention are: 
\vspace{-0.1cm}
\begin{equation}
    \textbf{\textit{Q}} = \textbf{\textit{X}}_{P}\textbf{\textit{W}}_{Q}, 
    \textbf{\textit{K}} = \textbf{\textit{X}}_{P}\textbf{\textit{W}}_{K}, 
    \textbf{\textit{V}} = \textbf{\textit{X}}_{P}\textbf{\textit{W}}_{V}
\end{equation}
\vspace{-0.5cm}
\begin{equation}
    Attention(Q,K,V)=Softmax(QK^T/\sqrt{D_k})V
    \end{equation}

where \textbf{\textit{Q, K, V}} are defined as the input of the self-attention, which are the query, key, and value, respectively. $\sqrt{D_k}$
 serves as a scaling factor to prevent SoftMax function from entering regions where it has extremely small gradients. The multi-head attention mechanism enhances the model’s ability to extract features by running several attention processes in parallel:
\begin{equation}
    MultiHead(Q,K,V) = Concat(head_1,...head_h)W^O
\end{equation}
\vspace{-1.1cm}

\begin{equation}
   Head_i = SelfAttention(QW^Q_i,KW^K_i,VW^W_i)
\end{equation}

where each head of ${Head_i}$ captures different aspects of the input, and Wo is the weight matrix for the output linear transformation that combines the outputs of all heads. During the decoding, encoded data and previously generated tokens are processed using self-attention to predict the next token. The final outputs are produced through a Softmax layer, mapping predictions to probabilities for tasks like text generation or classification \cite{12}.

The simplest way to use a Transformer in EEG processing is to apply the above backbone model directly. In this regard, the Encoder is predominantly used due to its ability to effectively capture the spatiotemporal characteristics of EEG, while the Decoder is less commonly employed, as tasks typically do not involve sequence generation. Overall, conventional temporal and spatial features formed by interactions between electrode channels, and spectral features are key EEG features. Studies have used the basic transformer encoder to extract these various features or their combinations. For instance, spatio-temporal sequence structures were designed to stack features \cite{13}, time-frequency branch structures were developed for feature fusion \cite{14}, and study \cite{15} applied different features to build separate models. Considering the classification based solely on partial dimensional features is often insufficient, models to combine spatio-temporal, label-label correlation features \cite{16}, and temporal-spatial-spectral features \cite{17,18}, are developed. Especially, AMDET proposed in \cite{19} converted EEG data into three-dimensional temporal-spatial-spectral representations, utilizing Transformer to extract spectral and temporal features, and thus enhancing emotion recognition. Apart from the feature extraction, Transformer model can also be used in applications of EEG source imaging (ESI) \cite{20}, semi-supervised domain adaption [21], and transfer learning using self-supervised pre-training methods \cite{22}. Moreover, similar to machine translation, the complete encoder-decoder structure has been attempted in Seq2seq tasks, such as the reconstruction of EEG-limb data \cite{23}.

\begin{figure}[htpb]
\centering
\includegraphics[width=8.3cm]{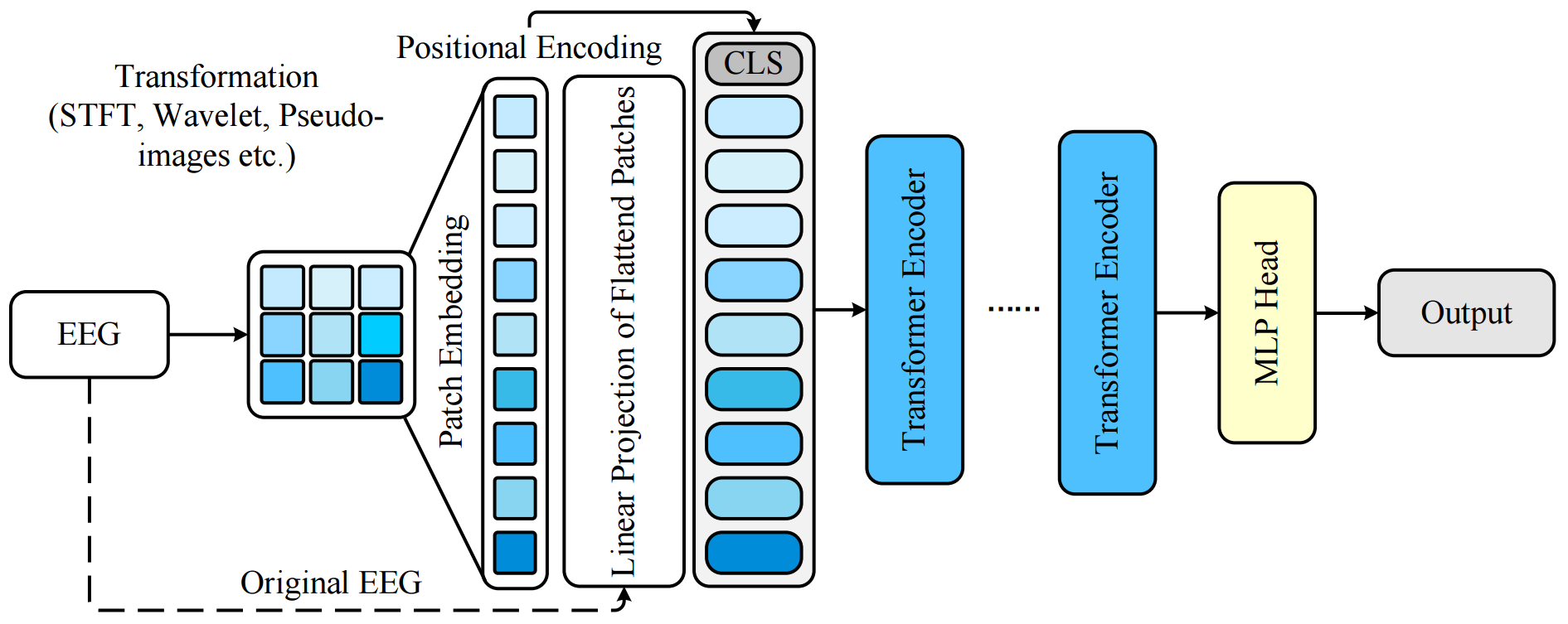}
\captionsetup{font=footnotesize}
\caption{General ViT backbone structure within EEG decoding.}
\label{fig 2}
\end{figure}

As a milestone applied in visual domain, Vision Transformer (ViT) has become a common specialized Transformer that processes images through a series of steps to achieve efficient classification \cite{24}. Since the EEG can be transformed into image-like formats, the general ViT backbone structure has been introduced in EEG decoding. As shown in \textbf{Fig.} \ref{fig 2}, during the Patch Embedding step, ViT divides the image into fixed-size patches and converts each patch into an embedding vector through linear projection. A [CLS] token is introduced for classification purposes. Position encodings are then added to each embedding vector to preserve the spatial relationships between patches. These embedding vectors are fed into the Encoder as a sequence, where the self-attention mechanism captures dependencies between image patches. Finally, the [CLS] token aggregates global features, which are then used by the classification head for image classification tasks.

The idea of converting EEG signals into images before processing them with the original ViT has been widely applied to various tasks of brain disease monitoring \cite{25}, sleep staging \cite{26,27}, and trust recognition \cite{28}, where image conversion techniques include short-time Fourier transform (STFT) \cite{25,29}, multimodal processing \cite{26}, and pseudo-images from raw signals \cite{28,30}. More specifically, in \cite{31}, the signals are divided into small segments according to the image patch segmentation method, while each channel of the signal in \cite{28} is treated as a patch and input into the ViT. Besides, ViT has also been used as part of ensemble learning \cite{32}, and as an independent classifier \cite{33}. In 2023, Li et al. \cite{34} proposed an automatic transformer neural architectures search framework to optimize ViT parameters for EEG-based emotion recognition. In 2024, a domain discrimination model based on ViT to extract common features from different datasets was designed \cite{35}, showing effectiveness in sensitivity, false alarm rate, and area under the receiver operating characteristic curve (AUC). 
\vspace{-0.2cm}

In conclusion, most of discussed methods rely on straight-forward architectures for specific tasks, with limited emphasis on the generalization and multi-dimensional feature extraction. Therefore, researchers are combining Transformer with other models and developing advanced variants. These achievements of a hybrid and customized models are detailed below.

\section{Hybrid Models}
\label{chap:3}
Although the Transformer model excels in extracting global dependencies in EEG, it has limitations in dealing with more detailed local features, especially short-term fluctuations and rapid changes. These local features, however, are critical for understanding the dynamics of brain activity. Moreover, the analysis of spatial features, for instance, to get connectivity patterns between different brain regions, also requires more fine-grained processing, while the basic transformer model may not adequately capture such information. Consequently, hybrid models of Transformer in conjunction with other DL-models have been widely investigated, which are concluded as follows.
\vspace{-0.4cm}
\subsection{CNN-Transformer}
\begin{figure}[htpb]
\centering
\includegraphics[width=8.8cm]{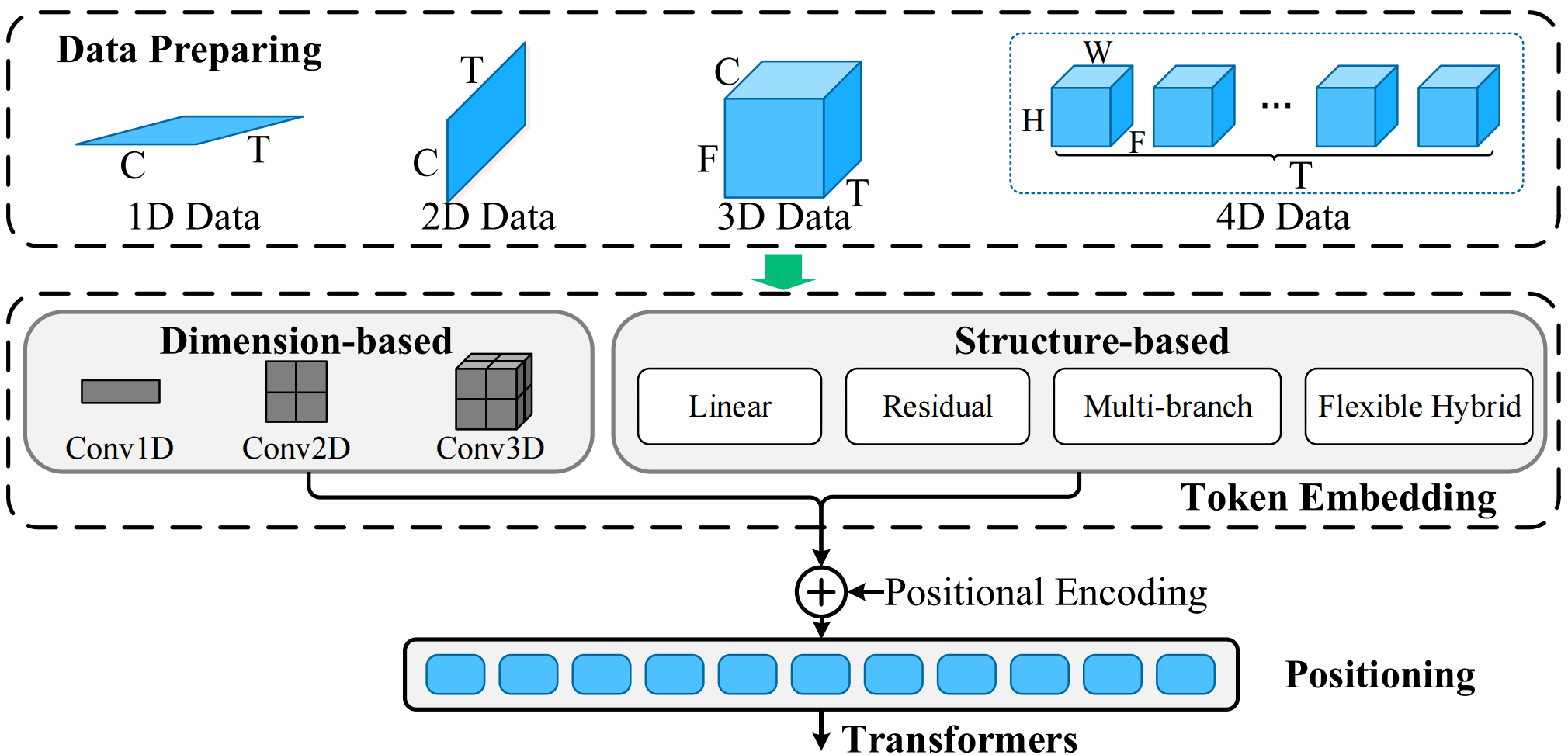}
\captionsetup{font=footnotesize}
\caption{General CNN-Transformer sequential architecture.}
\label{fig 3}
\end{figure}

Before adoption of Transformers, CNNs were the dominant models in the field of deep learning, and they remain widely used to date. However, CNNs have inherent limitations in handling sequential data \cite{36}, particularly for biological signals such as EEG. Such limitations stem from the convolution operation, which focuses mainly on local features and is not sufficiently sensitive to long-range dependencies within sequences. These issues, however, can be effectively addressed by Transformers, thus providing new perspectives. 
For a hybrid CNN-Transformer, CNNs are often exploited to perform the tokenization and embedding of EEG data, while Transformer extracts the entailed global features from the embedded representations. As shown in \textbf{Fig.} \ref{fig 3}, the relevant research advances are elaborated according to the CNNs design style and role in feature extraction, which are categorized as the dimension- and structure-based convolution models. 

1) \textit{CNN Dimension-based Hybrid Model:} One-dimensional convolution (Conv1D) is widely applied in sequential data analysis. The early hybrid CNN-Transformer model often used Conv1D first to extract temporal features locally, while the subsequent Transformer module integrated these features for more detailed analysis\cite{7,15,37,38,39,40,41,42,43,44,45,46}. The fixed kernel sizes of standard Conv1D limits its ability to capture both short- and long-term features. To overcome this, multi-resolution Conv1D was proposed to apply various kernel sizes in parallel within a single layer, where smaller kernels capture high-frequency, short-term details, and larger ones focus on the others. These extracted features are then fused via concatenation \cite{47,48}, weighted fusion \cite{49,50,51,52,53} or Transformer \cite{54,55,56,57,58}, to form a comprehensive multi-scale feature repre¬sentation. As for tasks requiring strict temporal consistency, causal convolution offers an alternative by ensuring each output depends only on current and past inputs, preventing information leakage from future time steps \cite{50,57}.

Except for the temporal features extracted by Transformer, Conv1D of the existing hybrid CNN-Transformer has also been effectively adapted for capturing additional EEG features, such as the spatio-frequency relationships between EEG channels \cite{59}, integrated temporal, spatial, and frequency features within single-channel data \cite{60}, and fused spatio-temporal informa¬tion \cite{15}. Conv1D also facilitates multi-branch fusion within multimodal applications \cite{61,62}. By leveraging Conv1D’s feature extraction capability, self-supervised learning further encodes data into segment embeddings for Transformers to predict missing information across spatiotemporal and Fourier domains \cite{40}, and improves EEG representation learning in generative \cite{37} and contrastive [39] frameworks. Building on these strengths, Conv1D plays a critical role in Foundation Models for large-scale EEG datasets \cite{45,63,64,65}.

To capture the spatial relationships between electrodes more efficiently, studies begin to adopt two-dimensional convolution (Conv2D). However, the direct Conv2D may yield limited results due to the inherently low spatial resolution \cite{66}. Therefore, scholars convert EEG into image-like formats first, which are then processed using Conv2D in raw, frequency, and time-frequency domains, with help of techniques of differential entropy (DE) \cite{67,68}, wavelet transforms \cite{70}, and power spectral density \cite{69,70}. Typically, a novel Conv2D based approach of \cite{71} involves integrating temporal, frequency, and time-frequency features into an image-like form for processing. For Conv2D operation, factorized convolution decomposes the square convolutions into two directional operations, which helps to improve computational efficiency. On the basis of such module, some research extracted temporal features alone, with Transformers \cite{72} or graph-based embeddings \cite{73} to capture spatial relationships. Examples of the independently applied factorized convolution for the spatiotemporal feature extraction can be referred from \cite{74,75,76,77,78,79,80,81}. Parallel factorized convolutions are attempted to extract temporal and spatial features while minimizing interference with original data \cite{15,82,83}. Another exemplified study of EEG-Net combines the factorized and depthwise separable convolutions to extract spatio-temporal-frequency features from single-channel 2D EEG data. Some studies also enhance EEG-Net by adding self-attention modules to capture global dependencies \cite{84,85}, and EEG Conformer \cite{86} combined factorized convolutions with self-attention mechanisms.

Additionally, Conv2D also supports multi-resolution analysis, enabling the extraction of features across temporal and spatial scales \cite{87,88,89,90}. For example, ADFCNN in \cite{76} used dual factorized branches to capture multiscale temporal and spatial features, while Ahn et al. \cite{87} integrated multiscale temporal, spatial convolution modules, and Transformer to create spatio-temporal-spectral features for EEG classification.

In summary, benefit from its adaptability and compatibility, Conv2D has become a cornerstone technique of the hybrid CNN-Transformers in handling both raw and transformed EEG data. However, the complexity of EEG’s high-dimensional features calls for even more advanced convolution operations for the feature extractor. Hence, three-dimensional convolution (Conv3D) appears to capture spatiotemporal and frequency features \cite{91,92}. In some cases, EEG data is even represented by four-dimensional formats of first three common dimension being the time dynamics, two-dimensional spatial distribution of electrodes, and another dimension being the channels \cite{93}, frequency characteristics over time \cite{42}, or the dynamic time-evolving features \cite{94}. These representations, though computationally intensive, enrich Transformer by leveraging EEG’s multidimensional features and offer a promising mean.

2) \textit{CNN Structure-based Hybrid Model:} According to the embedded CNN structure by the hybrid network, the developed architectures can be divided into the linear, multi-branch, residual, and flexible hybrid ones.

\begin{figure}[htpb]
\centering
\includegraphics[width=8.7cm]{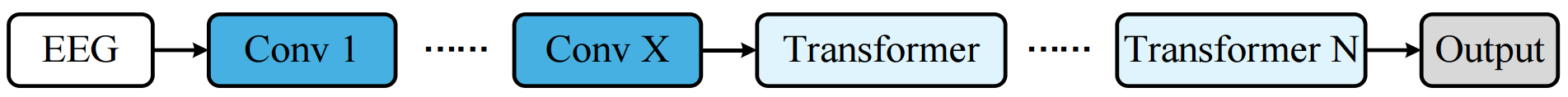}
\captionsetup{font=footnotesize}
\caption{General CNN-Transformer backbone with linear sequential structure.}
\label{fig 4}
\end{figure}

\textbf{\textit{Linear Structure}}: As shown in \textbf{Fig.} \ref{fig 4}, linear convolutional pathways often consist of straightforward, sequentially stacked convolutional layers, where each layer depends solely on the output of its predecessor. For EEG, linear structures typically rely on shallow Conv1D and Conv2D layers, optimized first to capture basic temporal and spatial features, and then be fed into Transformer to capture global dependencies of spatio-temporal dimensions \cite{15,42,95}. Such integration provides a simple and effective framework for the efficient feature extraction and representation, particularly for tasks requiring minimal architectural complexity. As decoding tasks grow more complex, more advanced CNN architectures emerged.

\begin{figure}[htpb]
\centering
\includegraphics[width=7cm]{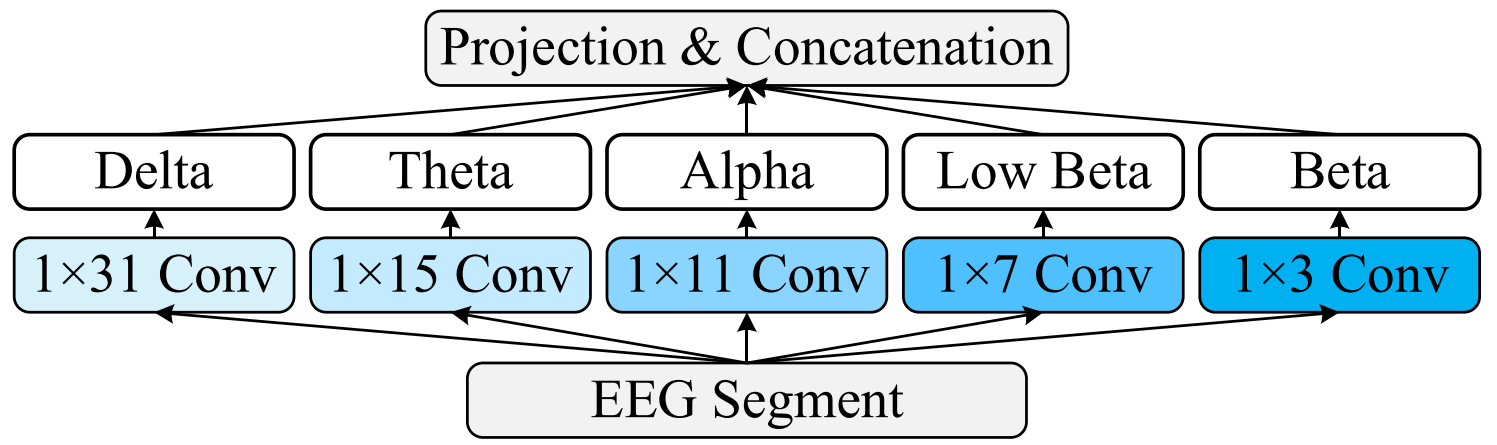}
\captionsetup{font=footnotesize}
\caption{One example of multi-branch structure-based CNN-Transformers.}
\label{fig 5}
\end{figure}

\textbf{\textit{Multi-Branch Structure}}: Multi-branch (parallel) convolution, popularized by GoogLeNet’s Inception module, addresses the limitations of single-scale kernels in linear structures. As a result, recent studies have introduced branched structures (e.g., parallel filters of varying sizes) to extract multi-scale features and use Transformer models for the feature fusion. In EEG decoding, to capture temporal features of different frequency bands, multi-branch designs are widely used by the two \cite{44,76,89,90}, three \cite{52,58,87,96}, four \cite{48,49,88}, and five \cite{47,54}. As an example, a five-branch model \cite{54}, as illustrated in \textbf{Fig.} \ref{fig 5}, isolates EEG bands using distinct filters, which are fused with a Transformer for feature integration.

\vspace{-0.2cm}

Beyond above multi-scale extraction, parallel convolutions also capture multi-dimensional features \cite{43,52,82,83,87}. Some studies arrange sequential factorized convolutions in parallel to combine features from the temporal, spatial, and frequency domains \cite{43,82}. These designs provide diverse feature inputs, enhancing CNNs’ effectiveness and comple-menting Transformer for robust EEG feature representation.

\begin{figure}[htpb]
\centering
\includegraphics[width=6.27cm]{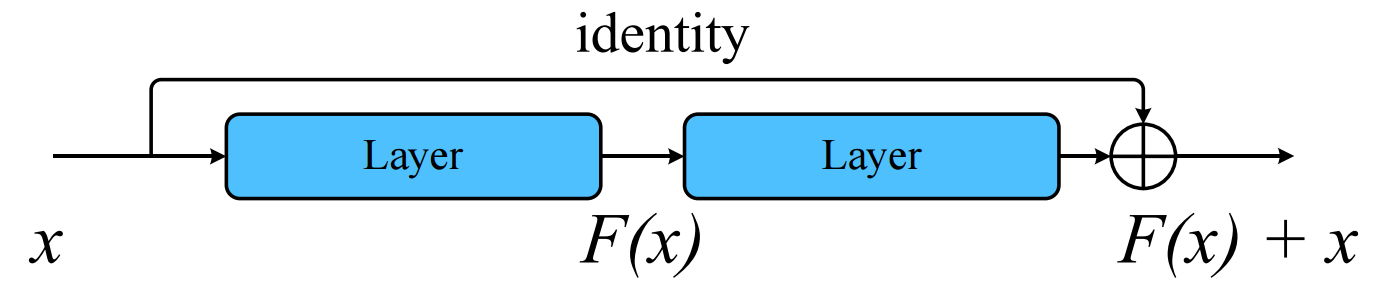}
\captionsetup{font=footnotesize}
\caption{A Res block that commonly used in hybrid CNN-Transformer.}
\label{fig 6}
\end{figure}

\textbf{\textit{Residual Structure}}: Residual structures like ResNet address vanishing gradients that allow inputs to skip directly over one or more convolutional layers, and to learn difference between the input and output. As shown in \textbf{Fig.} \ref{fig 6}, the input x is transformed by convolutional layers to produce \textit{F(x)}, which is then added to x via a shortcut connection. This structure enables the model to build very deep network architectures without performance degradation. Specifically for EEG decoding, such the module is adapted by hybrid CNN-Transformer to meet the unique demands \cite{43,48,50,51,97}. For example, \cite{97} employs ResNet-50 as an encoder to extract high-dimensional local features, which are then optimized by the self-attention module of Transformer, enhancing both feature precision and overall model performance. In summary, the structures mitigate gradient and degradation issues in deep networks, promoting information flow and stability of hybrid CNN-Transformers.

\textbf{\textit{Flexible Hybrid Architectures}}: Beyond above sequences, recent studies have expanded to more flexible designs, where the developed model in \cite{74} extracts global temporal features first, refining spatial details with separable convolutions. Study \cite{98} uses asymmetric branches where CNN and Transformer components target different feature scales. A Transformer for EEG and convolutional branch for EOG processing is fused via self-attention in \cite{62} for enhanced multimodal analysis.

To conclude, the different mentioned CNN structures in the hybrids have enhanced the feature integration across temporal, spatial, and frequency domains. Further, to finish the specific tasks, latest hybrid models of Transformer incorporating other advanced architectures are gaining attraction.
\vspace{-0.3cm}

\subsection{GAN-Transformer}
Generative adversarial networks (GANs) have found broad applications in data generation, image reconstruction, and domain adaptation, for which the ability to capture long-term dependencies of EEG can be enhanced by combining with Transformers. Sartipi and Cetin proposed a related method for emotional recognition (ER) task in a cross-subject manner \cite{99}, where Transformer was integrated with adversarial discrimi¬native domain Adaptation (ADDA) to improve EEG spatial features and minimize inter-subject differences. Results on the public DEAP dataset demonstrated the classification improve¬ments. Yu et al. \cite{100} designed a Transformer-based GAN to address class imbalance in sleep monitoring, and the results shown its potential for time series data generation tasks. Duan et al. \cite{101} used Transformers to extract features from EEG, and corresponding images were generated using conditional GAN (CGAN). Also in the field of image reconstruction, Zhao et al. \cite{102} introduced a Dual AxAtGAN method, which used the self-attention mechanism to replace traditional CNNs and enhance the model’s ability to capture long-range EEG dependencies. Combining GANs with Transformers holds great potential in EEG for improving feature extraction, synthetic data generation, and domain adaptation.

\subsection{Diffusion-Transformer}
Also emerging for generative tasks, diffusion models have known for their ability to reconstruct data through iterative denoising, producing stable and high-quality outputs, which has been integrated with the Transformer like DiT \cite{103}. In EEG field, diffusion-Transformer have also enabled significant progress. DiffMDD in \cite{104} combined diffusion models with Transformers to isolate noise-independent information for effective data augmentation. STADMs in \cite{105} improved EEG data quality by reconstructing low-resolution signals into super-resolution formats, and \cite{106} leveraged a DiT-based approach to generate high-quality synthetic EEG, addressing data scarcity challenges. A Transformer-based EEG encoder was used in a diffusion model in \cite{107} to enable EEG-to-image synthesis, paving the way for novel BCI applications. By integrating diffusion models with Transformers, EEG processing has gained another effective tool for robust data generation, super-resolution, and multimodal synthesis.
\subsection{GNN-Transformer}
Mapping EEG electrodes as nodes is one way to extract the complex EEG spatial features. Graph neural networks (GNNs) can model EEG signals as graphs, capturing the complex interactions between different brain regions to establish spatial features, which aids in a more comprehensive understanding of brain activity. For such innovative research ideas, Sun et al. \cite{108} introduced DBGC-ATFFNet-AFTL (DANet) framework, which combines dual-branch graph convolution, adaptive Transformer feature fusion, and adapter fine-tuning transfer learning. The method significantly improved the cross-subject emotion classification in accuracy and parameter efficiency. Similarly, EmoGT \cite{109}, used for ER task, integrates graph convolutional networks (GCN) with Transformer to optimize spatial relationship extraction and sequential information processing. This model serves as a core module for handling multimodal inputs and enhances recognition performance.

Kim et al. \cite{110} proposed the DGTM model, which includes a graph encoding module and a graph transformer module. This model was validated to accurately capture low-dimensional manifolds in EEG and the relationships between feature maps, improving data classification accuracy. Wang et al. \cite{111} developed a model that uses a multi-branch feature extractor to extract temporal, spatial, and frequency features. It employs multi-GCN to learn deep graph structures and utilizes a channel-weighted Transformer for feature fusion. In addition, SAG-CET of \cite{112} comprises a scale-aware adaptive GCN and a cross-EEG Transformer to capture multi-scale features and correlations, while CAW-MASA-STST (C-M-S) proposed in \cite{113} uses multilevel adaptive spectral aggregation (MASA) to capture the dynamic spectrogram of EEG by aggregating the spectral features of different sub-bands through graph convolution. In summary, through representing EEG electrodes as nodes in a graph, the hybrid GNN-Transformer framework facilitates a more comprehensive analysis of brain activity and enhances the extraction of spatial features.
\vspace{-0.4cm}
\subsection{RNN/LSTM-Transformer}
RNNs/LSTMs are effective for handling sequential data, while struggling with capturing long-term dependencies. Some studies reduce the receptive field of RNNs/LSTMs, using them as local feature extractors while using Transformers as global one \cite{114}. EEGAlzheimer’sNet developed in \cite{115} combines CNN, RNN, Transformer, and LSTM. Multi-scale dilated CNN and RNN extract spatial and temporal features, while an optimized Transformer-LSTM detects Alzheimer’s disease using weights obtained through the EWGLO algorithm. As a result, the developed model attained a satisfactory accuracy of 96\% and the Matthews Correlation Coefficient of 98\%. Pham et al. proposed a similar combined CNN-Transformer-LSTM model in \cite{116}, and Zhou et al. \cite{117} designed the temporal self-attention regression module in EmoTVR. In TBEEG \cite{118}, the relationships between different temporal dimensions were acquired by RNN-Transformer module with ViT combined to process frequency domain information, thus enhancing the model’s overall performance. Plus, BAFNet \cite{119} includes an LSTMTrans model that replaces the linear mapping in the Transformer encoder with LSTM. Despite these advances, given the computational complexity and potential model redundancy, it must overcome the high computation burden and potential efficiency issues in practice of combining RNN/LSTM and Transformer for EEG processing.
\subsection{SNN-Transformer}
Spiking neural networks (SNNs) mimic signal transmission mechanism of biological neurons \cite{120}, providing new insights into brain computational mechanisms. Study \cite{121} proposed the Spiking Conformer model, with the spiking convolution module processes information in both temporal and spatial dimensions, and spiking Transformer encoder utilizes spiking self-attention to capture the signal’s overall characteristics. There have also attempts to combine SNN with Transformer \cite{122}, which inherits the dynamic response characteristics of biological neurons and incorporates the global information processing advantages of Transformer.
\subsection{Capsule Network-Transformer}
Capsule network is a network architecture initially designed for processing image data and aims to address challenges faced by traditional CNNs \cite{123}. In EEG analysis, capsule networks capture the complex spatial relationships of EEG features with limited data. TC-Net \cite{124} encodes features with capsules and captures spatial relationships between multiple EEG features by a dynamic routing mechanism, while integrating the Trans¬former module to extract global information. MES-CTNet proposed in \cite{125} further enhances spatial attention, improving emotion state recognition accuracy through multi-dimensional feature input and various attention mechanisms. However, the application of capsule networks in EEG decoding tasks is limited by their complex structure, high computational cost, and need for improved robustness.
\section{Customized Transformers}
\label{chap:4}
While the hybrid approaches are well-established, ongoing research continues to explore more effective architectures. This section gives customized Multi/modified-encoder architectures, Pyramid, and Reconstructed Transformers.
\subsection{Multi Encoders}
\begin{figure}[htpb]
\centering
\includegraphics[width=8.8cm]{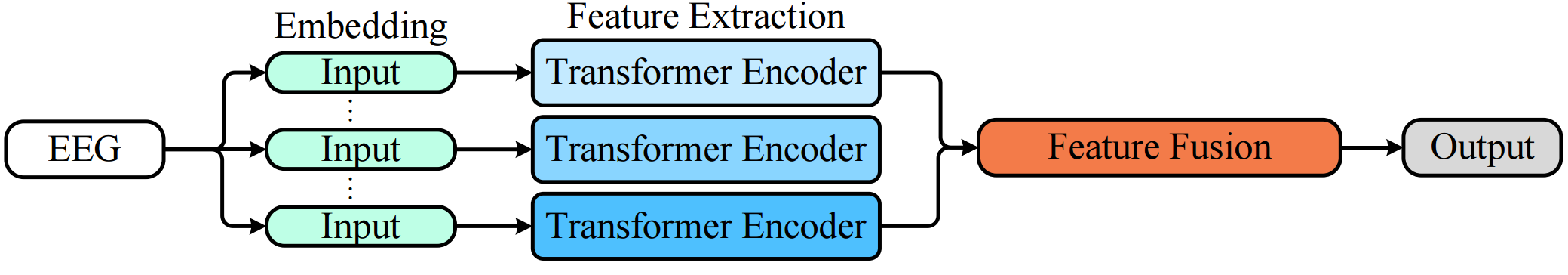}
\captionsetup{font=footnotesize}
\caption{General multi-encoder backbone structure of EEG decoding.}
\label{fig 7}
\end{figure}

The single encoder architecture often struggles to capture the complex features in EEG signals. To address this, researchers have developed multi-encoder architectures, where multiple Transformer encoders, as illustrated in \textbf{Fig.} \ref{fig 7}, process different feature branches in parallel, followed by a fusion mechanism. Based on the fusion approach, such the mentioned model can be divided into two categories.

\textit{1) Basic Multi-Encoders}: With straightforward fusion means of concatenation, weighted fusion, and attention mechanisms, the structures integrate the EEG features from diverse data sources or modalities of varied types, such as time-frequency and spatio-temporal characteristics from raw EEG \cite{126}, multi-scale temporal \cite{39,56}, spatial features \cite{90}, different data augmentation features \cite{127}, EEG and language features \cite{128}, transformed features from PSD \cite{56}, wavelet analysis \cite{69,129}, and linear-nonlinear features \cite{130}. Cross-modal features from datasets can also being integrated, such as EEG with EOG data \cite{131,132}, various types of sleep data \cite{61}, and audio spectrogram \cite{44}.

\begin{figure}[htpb]
\centering
\includegraphics[width=8.8cm]{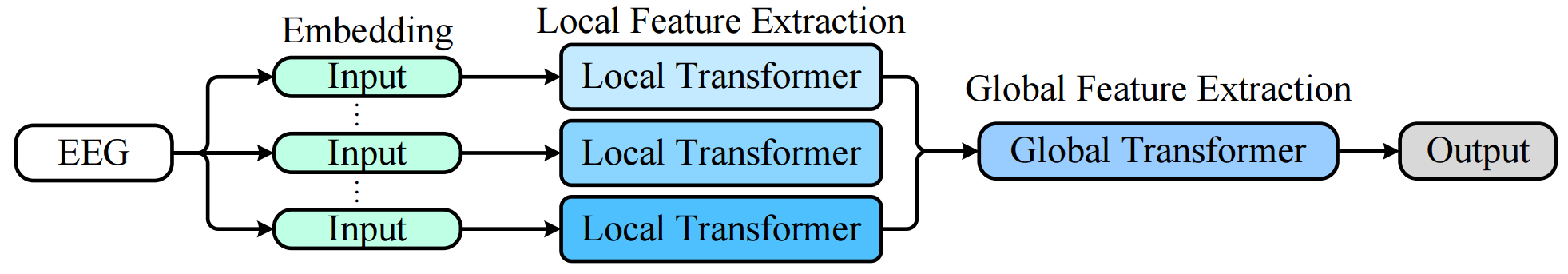}
\captionsetup{font=footnotesize}
\caption{General TNT-like backbone structure of EEG decoding.}
\label{fig 8}
\end{figure}

\textit{2) TNT \& TNT-like Model Application}: Theoretically, TNT \cite{133} extends the ViT with a dual-layer structure that captures both micro and macro image features. Similarly, the TNT-like model used for EEG decoding has been depicted in \textbf{Fig.} \ref{fig 8}, where multiple Transformer encoders are used to process local features and one global encoder integrates global features, thus making the model more effective (see examples of HSLT \cite{134}, SleepTransformer \cite{135}, VSTTN \cite{136}, EEG2Text \cite{137}). Kim et al. proposed SleepTNT \cite{138}, which captures both local features within different time periods and global features across the time periods, whereas the latter SleepConvTNT in \cite{139} further introduced a one-dimensional convolution Transformer module for automatic sleep staging.

Influenced by \cite{135}, similar models of SPDTransNet \cite{140}, hierarchical Transformer \cite{141}, and others \cite{142} have been proposed recently. Among them, the hierarchical Transformer consists of a high-level transformer (HLT) and a low-level transformer (LLT). The LLT extracts information within short time intervals, while the HLT captures global information across time intervals. MultiChannelSleep-Net in \cite{143} uses an encoder to extract information within time-frequency data converted from single-channel EEG. A unique TNT structure of B2-ViT captures generalized spatio-temporal correlations across encoder layers via a breadth Transformer layer \cite{144}.

Although these above multi encoder structures enhance EEG decoding via parallel processing and fusion, currently, their computational cost is still being relatively high. Hence, an idea to adjust the intrinsic encoder structure for customized Trans¬formers, which we called the modified encoders has aroused.

\subsection{Modified Encoders}

\begin{figure}[htpb]
\centering
\includegraphics[width=8.5cm]{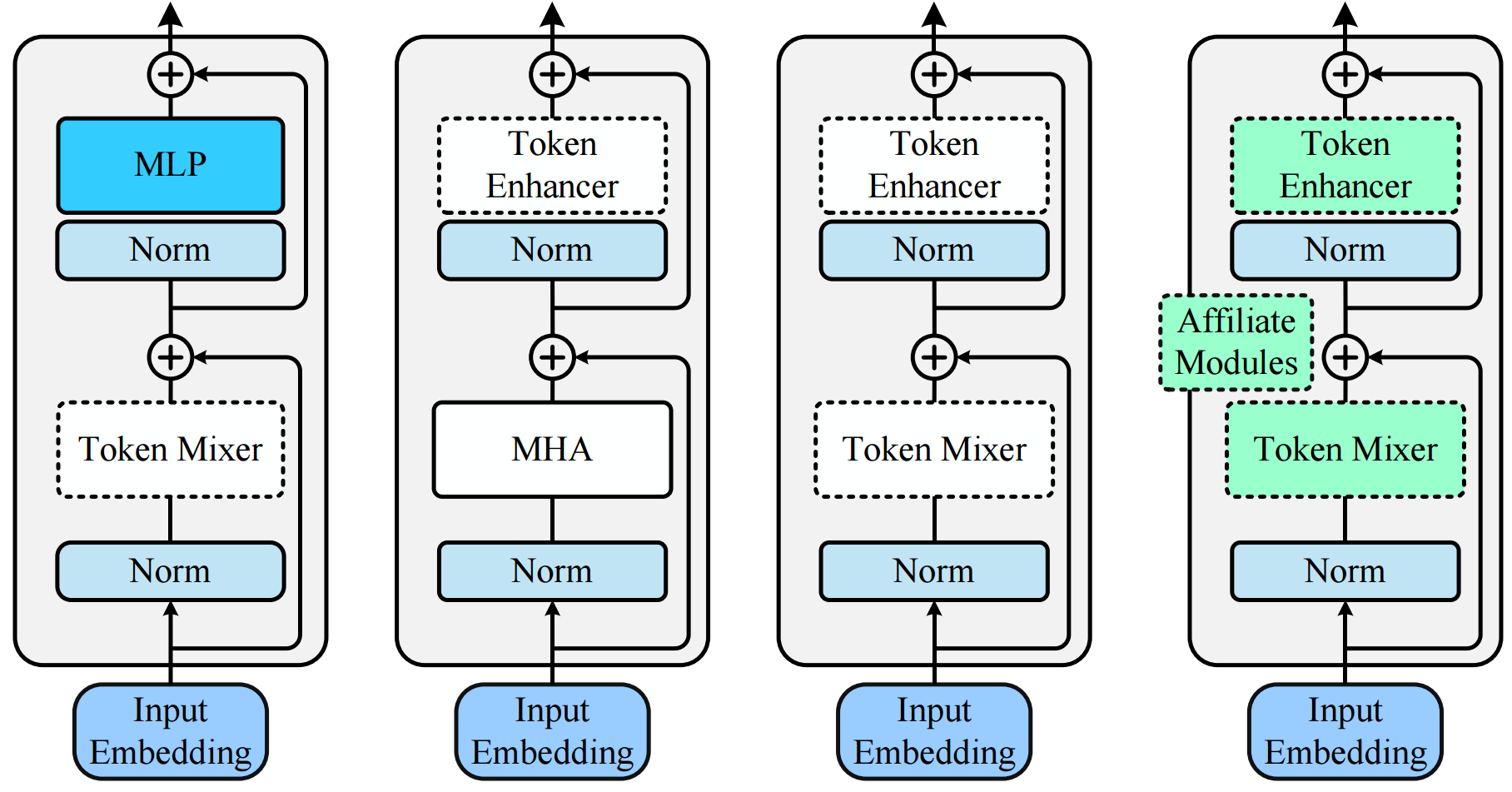}
\captionsetup{font=footnotesize}
\caption{Evolution of modified encoder frameworks based on MetaFormer, proposed in this work, to enhance adaptability for EEG decoding tasks. From the left to right are structures of the Token Mixer, Token Enhancer, Extended MetaFormer (EM), and EMA frameworks (i.e., EM with Affiliate Modules).}
\label{fig 9}
\end{figure}

A generalized framework of MetaFormer from prior research \cite{145} redefines the multi-head attention as a “Token Mixer”, emphasizing that the effectiveness of Transformers lies in their structural design rather than specific mechanisms. Expanding on this principle, as shown in \textbf{Fig.} \ref{fig 9}, current studies with four modified encoder structures have been concluded as follows.

\textit{1) Token Mixer-based Encoders}: Such implementations generally fall into two categories, where the first focuses on enhancing the Multi-Head Attention by introducing new mechanisms to improve performance or replacing MHA with alternative modules to reduce computational and memory demands, particularly for long-sequence EEG. Examples are Retention \cite{146}, Sparse Attention \cite{147,148}, One-Way Self Attention \cite{149}, Linear Self Attention \cite{150}, 2D AvgPooling \cite{151}, and Spiking Self Attention \cite{121}. For these methods, techniques of low-rank approximations, sparse connections, directional constraints, and biologically inspired designs have been leveraged. Composite attention of recent studies also enhances feature representation by integrating multiple attention mechanisms. Study \cite{152} developed shifted channel attention for the global and local feature extraction. ESSleep \cite{153} stacks two encoders to extract features at epoch and sequence levels, and \cite{154} adopts a parallel approach, running two encoders simultaneously for spatiotemporal feature extraction. Moreover, diverse attention types within a single encoder has been attempted in \cite{94,155}. Typically, \cite{156} combines intra- and inter-MHA for multi-scale channel dependencies. These mechanisms effectively capturing EEG’s complex multi-dimensional dependencies.

The second category, on the other hand, stresses multi-modal and multi-scale feature extraction. Specifically, features at different scales are targeted by Channel Attention (CA) \cite{157} and Shifted Channel Attention \cite{152}, and mechanisms like TcT Attention \cite{158} and TSA/SSA \cite{159} specialize in extracting spatio-temporal relationships. In multi-modal scenarios, TACO Cross Attention supports seamless integration of diverse data modalities \cite{160}. These Token Mixer based encoders has effectively enhance the flexibility of Transformers, enabling robust processing of long-sequence and multi-modal EEG data.

\textit{2) Token Enhancer-based Encoders:} Compared to MHA, the multilayer perceptron (MLP) is responsible for feature expan¬sion and transformation, and can be adapted to better capture multi-scale patterns in EEG data. A multi-branch feed-forward network (FFN) structure processes wavelet features across frequency bands, facilitating knowledge transfer between raw and transformed EEG data \cite{161}. Similarly, the Convolutional Feature Expansion (CFE) module substitutes MLP with multi-scale convolutions, extracting complex temporal and spatial dependencies \cite{89}. These enhancements to the Token Enhancer can improve EEG decoding performance while preserving the core self-attention mechanism.

\textit{3) Extended MetaFormer (EM)-based Encoders:} Such the customized framework simultaneously replaces both the MHA and MLP, offering greater flexibility for EEG decoding. More specifically, MHA adaptations primarily aim to enhance the computational efficiency and scalability, such as the Separable MHA \cite{55}, Prob-Sparse Attention \cite{162}, and Squeezed MSA \cite{163,164}, which utilize sparse connections and low-rank approximations. CNN-based Token Mixers \cite{165}, as employed in SSVEPFormer, enable channel-specific feature extraction, showcasing adaptability for EEG tasks. The MLP substitutions focus on improving feature extraction and reducing computa¬tional costs. Convolution-based FFNs of BFFN \cite{55}, RFFN \cite{163,164}, are common approaches. Besides, Multi-branch expert structures \cite{162} and Channel MLP \cite{165} further enhance multi-scale and inter-channel feature learning.

\textit{4) EM with Affiliate Modules (EMA):} Affiliate Modules are additional components integrated into Transformer structure to expand its functionality, including gating mechanisms \cite{166}, positional feedforward mechanisms \cite{167}. Convolutions remain a key support module, for which causal convolution accelerates time-series processing \cite{50,57}, and standard convolution extracts local feature relationships \cite{55,77,164,167}. Some modules also combine convolution with pooling layers \cite{77} to form distillation modules \cite{147} that preserve key features. Together, Affiliate Modules within EM architecture enhance the decoding adaptability and flexibility.

\begin{table*}[htbp]
   
    \captionsetup{font=footnotesize,justification=centering}
    \caption{\textsc{\\Overview Of Reconstructed Architecture Transformer [BD: Brain Disorders; SP: Seizure Prediction; SSC: Sleep Stage Classification; ACC: Accuracy; RRMSE: Relative Root Mean Squared Error; CC: Correlation Coefficient; \text{ $\kappa$}: Cohen Kappa; AUC: Area Under The Curve; SEN: Sensitivity; MF1: Macro-Average F1-Score; TPR: True Positive Rate; FPR: False Prediction Rate; CID: Cognitive Impairment Detection]}}
     \label{table1}
    \centering
    \footnotesize
    \resizebox{\textwidth}{!}{
    \begin{tabular}{
    >{\centering\arraybackslash}m{1.10cm}>{\centering\arraybackslash} m{0.75cm} >{\centering\arraybackslash}m{0.5cm} >{\centering\arraybackslash}m{1cm} >{\centering\arraybackslash}m{3.55cm} >{\centering\arraybackslash}m{3.65cm} >{\centering\arraybackslash}m{1.75cm} >{\centering\arraybackslash}m{6.0cm}}
        \toprule
    \makecell[c]{\textbf{Year}} & 
    \makecell[c]{\textbf{Models}} & 
    \makecell[c]{\textbf{Ref}} & 
    \makecell[c]{\textbf{Tasks}} & 
    \makecell[c]{\textbf{Model Characteristics}} & 
    \makecell[c]{\textbf{Structure}} & 
    \makecell[c]{\textbf{Dataset}} & 
    \makecell[c]{\textbf{Evaluation}} \\
        \midrule
        \multirow{3}{*}{MSDTT} & \multirow{3}{*}{2022} & \multirow{3}{*}{\cite{70}} & \multirow{3}{*}{ER} & \multirow{3}{*}{\makecell{Extract spatial features and \\ temporal features simultaneously}} & \multirow{3}{*}{\makecell{Multi-spatial Transformer \& \\  Dynamic Temporal Transformer}} &  SEED & Acc 97.52\% \\
         & & & & & &  SEED-IV & Acc 96.70\% \\
         & & & & & &  DEAP & Acc 98.91\% \\
         \multirow{3}{*}{\makecell{Denosiefo\\rmer}} & \multirow{3}{*}{2023} & \multirow{3}{*}{\cite{53}} & \multirow{3}{*}{Seq2seq} & \multirow{3}{*}{\makecell{Excellent local and long-term \\ feature representation capabilities}} & \multirow{3}{*}{\makecell{Residual Information Enhanced \\ Encoder-Decoder Structure} } &  EEG-EOG & RRMSE 0.278, CC 0.951 \\
         & & & & & &  EEG-EMG & RRMSE 0.389, CC 0.894 \\
         & & & & & &  Real-Labo & Acc 98.91\% \\
         \multirow{2}{*}{GAT} & \multirow{2}{*}{2023} & \multirow{2}{*}{\cite{83}} & \multirow{2}{*}{MI} & \multirow{2}{*}{Cross Subject, Domain Adaptation} & \multirow{2}{*}{\makecell{Global Adapter, \\ Temporal Attention} } &  BCI-IV 2a & Acc 76.58\%, \text{ $\kappa$} 0.6877 \\
         & & & & & &  BCI-IV 2b & Acc 84.44\%, \text{$\kappa$} 0.6889 \\
         \multirow{2}{*}{M-d-C} & \multirow{2}{*}{2023} & \multirow{2}{*}{\cite{111}} & \multirow{2}{*}{BD(SP)} & \multirow{2}{*}{Graph Conv, Multi-Branch/view} & \multirow{2}{*}{\makecell{MB Feature Extractor, \\dMGCN, CWTFFNet }} &  CHB-MIT & AUC 93.5\%, SEN 97.8\%, FPR 0.059 \\
         & & & & & &  Xunwu PD & AUC 98.4\%, SEN 100\%, FPR 0.079 \\
          
         \multirow{2}{*}{C-M-S} & \multirow{2}{*}{2023} & \multirow{2}{*}{\cite{113}} & \multirow{2}{*}{EEG Visual} & \multirow{2}{*}{Dual-Branch, GCN} & \multirow{2}{*}{CAW, MASA, STST} &  EEG72 & Acc 54.82\%/29.98\% of 6/72-class \\
         & & & & & &  EEG200 & Acc 68.33\%/31.37\% of 2/10-class \\
         \multirow{6}{*}{BIOT} & \multirow{6}{*}{2023} & \multirow{6}{*}{\cite{150}} & \multirow{6}{*}{Seq2seq} & \multirow{6}{*}{\makecell{Bio-signal to “bio-signal \\sentences”}} & \multirow{6}{*}{\makecell{Bio-signal Tokenization, Linear \\ Transformer Encoding}} & CHB-MIT & Acc 66.40\%, AUC-PR 0.26, AUROC 0.86 \\
         & & & & & &  IIIC Seizure & Acc 57.62\%, \text{ $\kappa$} 0.4932, Weighted F1 0.577\\
         & & & & & &  TUAB & Acc 79.25\%, AUC-PR 0.87, AUROC 0.87 \\
         & & & & & &  TUEV & Acc 46.82\%, \text{ $\kappa$} 0.4482, Weighted F1 0.709 \\
         & & & & & &  PTB-XL & Acc 83.15\%, AUC-PR 0.90, AUROC 0.75 \\
         & & & & & &  HAR & Acc 94.61\%, \text{ $\kappa$} 0.9351, Weighted F1 0.946 \\
          \multirow{2}{*}{C-T-D} & \multirow{2}{*}{2024} & \multirow{2}{*}{\cite{52}} & \multirow{2}{*}{RSVP} & \multirow{2}{*}{Domain-Rectified framework} & \multirow{2}{*}{CST, TVA, DRTL frame} &  Tsinghua RSVP & Acc 92.56\%, TPR 89.62\%, AUC 0.9415, \\
         & & & & & &  PhysioNet RSVP & Acc 72.02\%, TPR 75.13\%, AUC 0.7281\\
         \multirow{3}{*}{TSP} & \multirow{3}{*}{2024} & \multirow{3}{*}{\cite{58}} & \multirow{3}{*}{ER} & \multirow{2}{*}{Temporal \& Spatial prediction} & \multirow{2}{*}{\makecell{Spatial Masked Autoencoder \& \\ BIOT}} &  SEED & Acc 66.53\%, \text{ $\kappa$} 0.4498 \\
         & & & & & &  SEED-IV & Acc 46.40\%, \text{ $\kappa$} 0.2737\\
         & & & & & &  TUEV & Acc 53.37\%, \text{ $\kappa$} 0.5261\\
         \multirow{2}{*}{-} & \multirow{2}{*}{2024} & \multirow{2}{*}{\cite{96}} & \multirow{2}{*}{ER} & \multirow{2}{*}{Dual-Branch, Multi-modal} & \multirow{2}{*}{\makecell{Multi-scale TASLM, Multi-view \\AECLM}} &  SEED & Acc 89.21\%, \\
         & & & & & &  DEAP & Acc 93.56\%, 93.53\%\\
         \multirow{3}{*}{\makecell{Tsformer-\\SA}} & \multirow{3}{*}{2024} & \multirow{3}{*}{\cite{174}} & \multirow{3}{*}{RSVP} & \multirow{3}{*}{Multi-view feature fusion} & \multirow{3}{*}{\makecell{Cross-view interaction module, \\ attention-based fusion module, \\ subject-specific adapter}} &  PD Task 1 & Acc 90.29\%, TPR 89.21\%, FPR 8.64\% \\
         & & & & & &  PD Task 2 & Acc 88.42\%, TPR 87.24\%, FPR 10.39\%\\
         & & & & & &  PD Task 3 & Acc 90.20\%, TPR 89.36\%, FPR 8.97\%\\
          \multirow{3}{*}{ECO-FET} & \multirow{3}{*}{2024} & \multirow{3}{*}{\cite{175}} & \multirow{3}{*}{ER} & \multirow{3}{*}{\makecell{Multi-modal, critical subnetwork \\ Selection SSL}} & \multirow{3}{*}{\makecell{Functional Emotion Transformer\\, critical subnetwork  selection, \\functional  brain connectivity }} &  SEED & Acc 93.69\% \\
         & & & & & &  SEED-IV & Acc 87.76\%\\
         & & & & & &  SEED-V  & Acc 77.13\%\\
         \multirow{2}{*}{MCAF-T} & \multirow{2}{*}{2024} & \multirow{2}{*}{\cite{176}} & \multirow{2}{*}{ER} & \multirow{2}{*}{Multi-modal} & \multirow{2}{*}{\makecell{Temporal Transformer, \\ Channel-wise Attention}} &  SEED & Acc 89.21\%, \\
         & & & & & &  DEAP & Acc 93.56\%, 93.53\%\\
          CoRe-Sleep & 2024 & \cite{177} & SSC & Multi-modal & EEG/EOG Encoder, Multimodal EEG/EOG Encoder &  SHHS & Acc 84.0\%, \text{ $\kappa$} 0.771, MF1 73.0\%\\
          BMMTrans & 2024 & \cite{178} &  CID &  Dynamic brain network structures in Riemannian space, integrating tangent mapping with self-attention &Brain Network sequence-driven Manifold-based Transformer &  PD & Acc 90\%, AUROC 0.91, Precision 0.87, Recall 0.92, F1 0.89\\
        \bottomrule
    \end{tabular}
    }
\end{table*}

Some representative examples fall in above categories are concluded in \textbf{Table A-I} (see supplements), with tasks, model characteristics, and compared evaluations results in detail.
\vspace{-0.4cm}
\subsection{Pyramid Transformer}
\begin{figure}[htpb]
\centering
\includegraphics[width=8cm]{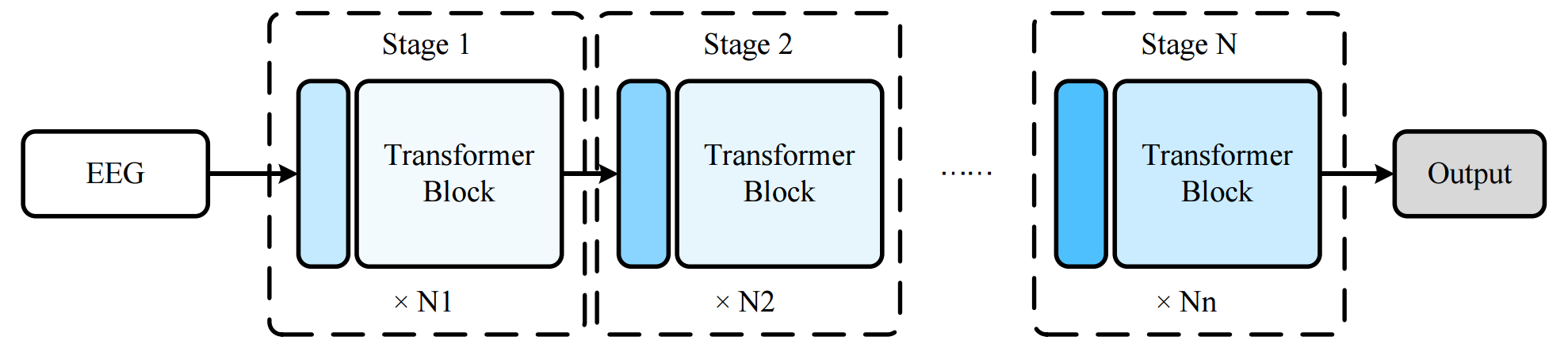}
\captionsetup{font=footnotesize}
\caption{General structure of Pyramid Transformer for EEG decoding.}
\label{fig 10}
\end{figure}

To reduce computational complexity while ensuring the high performance, pyramid Transformer was inspired by CNN feature pyramids to build a hierarchical framework by applying different techniques for down-sampling and up-sampling the activations within the Transformer (see \textbf{Fig.} \ref{fig 10}).  The structures are showcasing their potential for advanced EEG processing tasks, and can be further categorized in two classes below.

1) PVT-like Structure: Pyramid Vision Transformer (PVT) \cite{168} employs spatial reduction attention to efficiently handle large-scale inputs by combining reduced resolutions with higher feature dimensions. PVT-like Transformers mainly process diverse input formats, such as raw signals \cite{77,163}, pseudo-images \cite{152}, etc. Specifically, ScatterFormer \cite{169} uses frequency-aware attention to extract features across frequency layers from wavelet-transformed data. JDAT \cite{170} integrates spatial-spectral-projection blocks with transformer encoders, enabling automatic multi-dimensional focusing in multispectral images. These examples demonstrate the adaptability of PVT-like architectures in EEG decoding.

2) Swin Transformer Structure: Building on PVT, the Swin Transformer \cite{171} uses shifted window attention, staggering local windows across layers to expand the receptive field while optimizing computational efficiency. For EEG processing, a channel attention mechanism has been incorporated to extract intrinsic correlations between channels \cite{172}, while \cite{173} constructed four-dimensional data features at the preprocessing phase and integrated spatiotemporal attention modules into Swin Transformer backbone. In conclusion, these Swin-based models represent a significant advancement in Transformer, demonstrating their ability to effectively manage the complex spatiotemporal relationships inherent in long-sequence EEG.

\subsection{Reconstructed Transformer Architecture}

Some models mimic the Transformer structure and undergo reconstruction to extract specific EEG features. \textbf{Table \ref{table1}} lists the related studies. More specifically, MSDTTs \cite{70} combine multi-domain spatial branch with dynamic temporal branch of Transformers to extract and weight corresponding features. Multimodal models of \cite{96} utilize a multi-scale temporal asymmetric learning module to learn brain characteristics from raw signals, and use another branch to extract facial emotion images related to EEG via a multi-view attention-enhanced convolutional learning module. Further, a cross-attention Transformer is applied to fuse multi-modal information, and extract EEG temporal-spectral-spatial features. GAT \cite{83} uses self-attention mechanisms and parallel convolution layers to capture spatiotemporal information, primarily designed for EEG decoding. Similarly, models like TSformer-SA \cite{174} adopt domain adaptation strategies to handle multimodal data.

ECO-FET \cite{175} is another multimodal model that enhances learning from eye movement data by combining functional brain connectivity with spectral-spatial-temporal domains. Studies \cite{176,177},  also use EOG-EEG data. BNMTrans \cite{178} extracts features from sequential brain networks using a self-attention mechanism and leverages geometric correlations within the Riemannian manifold to guide feature extraction, optimizing cognitive impairment monitoring. TSP \cite{58} uses spatial masking autoencoders and temporal contrastive predictive coding to extract diverse information from EEG. BIOT in \cite{150} focuses on converting different biological signals into a unified representation. \cite{53} combines a residual variational architecture with a slice mode attention mechanism to construct Transformer-based encoder-decoder models to remove EEG artifacts. Besides, C-M-S \cite{113} decodes semantic information of visual stimuli from a single EEG signal, and significantly improves decoding accuracy.

To summary, by introducing specific changes to the original Transformer architecture, such as subtle structural adjustments of encoders, or novel modifications to overall networks, the model’s ability to capture temporal, spatial, and spectral features has been progressively enhanced.

\section{Discussion}
\label{chap:5}
The Transformer architecture has garnered significant interest for its exceptional ability to handle sequential data, for which related models can not only be seamlessly integrated into other deep learning networks but can also be customized with various adaptations in EEG decoding. For these Transformer-based networks, we have released detailed summarization of current available open-source EEG datasets in Table B-I (see in supplementary materials), which includes the data size, tasks,
annotations, number of subjects, and related studies. Among them, datasets with different data size have been attempted. For those data-scarce situations, extra auxiliary data are integrated and unsupervised or self-supervised learning methods are used by leveraging unlabeled data \cite{179}. Moreover, current Transformer-based approaches are nearly all essentially devoted to better synergize the temporal, spatial, and frequency features of EEG. We conclude the specific extracted features of models in supplemental Table C-I, from which we summarize some particular strengths of each model. Specifically, since EEG signal possesses good temporal properties, the relevant temporal features have been exploited by most of current studies, except for \cite{15,25,28,34,68,148,165,169}. Meanwhile, various Transformer-based networks are concentrating on the complex spatial relationships among multiple channels of electrodes. Interestingly, current Trans¬formers, no matter in which structures, have limited applications to separately dig into frequency domain features. Instead, most studies converted EEG into time-frequency data or combine them with other features for processing. In particular, the temporal-spatial-spectral representations of EEG have been simultaneously investigated by categorized models of backbone \cite{17}, hybrid models (with CNN \cite{40,85}, GAN \cite{101}, RNN \cite{118}, CapsNet \cite{123,124}), and customized ones (e.g., multi-Encoder \cite{126,143}, modified Encoder \cite{77,157}, Pyramid \cite{170,172}, reconstructed \cite{52,83}).

Overall, most of these mentioned Transformer-based models integrate EEG patterns from multi-sources and multi-features well, which enhances the feature extraction capability and exhibits certain robustness. However, the scarcity of available data, the limitations of generalization ability, computational resources challenges, and weak interpretability continue to constrain their wide application. Here we specifically discuss these challenges and possible directions for future research breakthroughs.

\begin{table*}[t!]
    \captionsetup{font=footnotesize,justification=centering}
    \caption{\textsc{\\Performance Comparison Between Models on Different Tasks and Datasets [STD: Standard Deviation; SPEC: Specificity; PR: Precision]}}
    \centering
    \footnotesize
    \resizebox{\textwidth}{!}{
    \begin{tabular}{>{\centering\arraybackslash}m{1.5cm} >{\centering\arraybackslash}m{1.5cm} >{\centering\arraybackslash}m{2.5cm} >{\centering\arraybackslash}m{2.5cm} >{\centering\arraybackslash}m{9cm}}
        \toprule
        \textbf{Task} & \textbf{Dataset} & \textbf{Model} & \textbf{Model Category} & \textbf{Performance} \\
        \midrule
        \multirow{5}{*}{\makecell{Emotional \\Recognition}} & \multirow{5}{*}{SEED} & DGCNN & CNN & Acc ± Std: 90.40 ± 8.49\% \\
        & & AMDET\cite{19} & CNN-Transformer & Acc ± Std: 97.17±0.93 \% \\
        & & ACTNN\cite{42} & CNN-Transformer & Acc ± Std: 98.47±1.72 \% \\
        & & DANet\cite{108} & GNN-Transformer & Acc ± Std: 98.47±1.72 \% \\
        & & MV-SSTMA\cite{57} & Modified Encoders & Acc ± Std: 98.47±1.72 \% \\
        \midrule
         \multirow{6}{*}{\makecell{MI \\Classification}} & \multirow{6}{*}{BCI IV 2a} & EEGNet & CNN & Acc: 74.50 \%, \text{$\kappa$}: 0.6600 \\
        & & EEG Conformer\cite{86} & CNN-Transformer & Acc: 74.50 \%\text{ $\kappa$}: 0.6600 \\
        & & ADFCNN\cite{76} & CNN-Transformer & Acc: 79.39\% \\
        & & MI-CAT\cite{78} & Modified Encoders & Acc: 76.74, \%\text{ $\kappa$}: 0.6900 \\
        & & GAT\cite{83} & Reconstructed &Acc: 74.50, \%\text{ $\kappa$}: 0.6877\\
        & & Dual-TSST\cite{81} & CNN-Transformer &Acc: 80.67, \%\text{ $\kappa$}: 0.7413\\
        \midrule
         \multirow{6}{*}{\makecell{Sleep \\Stage \\Classification}} & \multirow{6}{*}{Sleep-EDFx} & DeepSleepNet & CNN & Acc: 77.80 \%, Macro F1 score: 0.7180, \text{ $\kappa$}: 0.7000 \\
        & & AttnSleep\cite{50} & CNN-Transformer & Acc: 84.20 \%, Macro F1 score: 0.7530, \text{ $\kappa$}: 0.7800 \\
        & & ResTrans\cite{43} & CNN-Transformer & Acc: 80.70 \%, Macro F1 score: 0.7410, \text{$\kappa$}: 0.7400 \\
        & & TSA-Net\cite{51} & CNN-Transformer & Acc: 82.21 \%, Macro F1 Score: 0.7351, \text{$\kappa$}: 0.7501 \\
        & & SleepTrans\cite{135} & Reconstructed &Acc: 84.9 \%, Macro F1 score: 0.7880, \text{$\kappa$}: 0.8280, SPEC: 95.9 \% \\
        & & FoME\cite{65} & Foundation Models &Acc: 99.03 \%, Macro F1 score: 0.9910, PR: 98.68\%, AUC: 98.68\%, SENS: 98.73\%, F1 score: 0.9910\\
      \midrule
       \multirow{6}{*}{\makecell{Brain \\Disorders}} & \multirow{6}{*}{CHB-MIT} & CNN & CNN & SENS: 81.20\%, FPR: 0.16\/h \\
        & & TTT\cite{17} & Transformer & SPEC: 96.23 \%, SENS: 96.01\%, FPR: 0.047/h \\
        & & MdC\cite{111} & GNN-Transformer & Acc: 79.39\% \\
        & & HviT-DUL\cite{55} & Modified Encoders & AUC: 93.5 \%, SENS: 97.80 \%, FPR: 0.059/h \\
        & & B2-ViT\cite{144} & Multi-Encoders &PR :92.30 \%, SENS: 93.33\%, FPR: 0.057/h\\
        & & TGCNN\cite{164} & Modified Encoders &AUC: 93.5 \%, SENS: 91.50 \%, FPR: 0.145/h\\
        \bottomrule
    \end{tabular}
    }
\end{table*}

\vspace{-0.4cm}
\subsection{Challenges of the Current Models}
\textit{1) The Scarcity of Available Data:} The first challenge lies in the data deficiency for effective training. Compared to traditional machine learning methods and simple convolutional neural networks, although Transformer-based models have unique advantages in handling sequence tasks, they typically require large amounts of training data. However, compared to the image or text required by foundation models, the collection of brain signals is rather more difficult \cite{180}. Despite the use of non-invasive means can avoid surgical risks, the acquisition of large applicable datasets has been a major challenge in the field of EEG decoding. The main reasons are as follows. To begin with, some existing datasets are not publicly available, creating difficulties in data sharing of the field. Second, the collection of EEG data requires the relevant ethical approval, the recruitment of users/subjects, and obtaining their consent forms, which is a time-consuming and laborious task. In particular, some EEG paradigms also require extensive training or stimulus assistance, during which users may interrupt the experiments due to discomfort. Besides, variations in sampling devices, users’ state, environment, time of day, and other factors may make EEG data unsuitable for the given decoding task. All these above pose additional challenges of data preparation. 
Additionally, the annotation of EEG also influences the datasets quality. For instance, the annotation of some emotion recognition task needs the combination of self-assessments with scales and physiological data. Clinical annotations, such as in sleep monitoring, are expert-driven, time-intensive, and prone to subjectivity. Thus, although there are some automatic annotation software, the biases among experts in the same task (e.g., sleep staging) undermines dataset reliability. 

\textit{2) The Limitation of Generalization Ability:} Despite the flexibility and adaptability of Transformer models allow them to be applied in a variety of ways, it is not yet possible to provide one-size-fits-all solutions to each specific EEG-related issue. As reported in \textbf{Table \ref{table1}}, and \textbf{Table A-I}, many existing models are designed for specific tasks and validated on limited datasets, which restricts their ability to generalize across different tasks. 

More specifically, the performance comparisons of specific models, with benchmark of CNNs, are concluded in Table II, and we can summarize that current Transformer-based models exhibit some task preferences, although not particularly pronounced. Overall, models have demonstrated robust generalization capabilities in tasks of emotion recognition on SEED dataset, motor imagery classification on BCIIV 2a dataset, sleep stage classification on Sleep-EDFx dataset, brain disorders diagnosis on CHB-MIT dataset, etc. These findings provide preliminary evidence of their ability to generalize across datasets and tasks. Nevertheless, further systematic evaluations across a broader range of datasets and experimental conditions are needed to substantiate these observations. Meanwhile, from Table II, we see that CNN-Transformer performs well in above mentioned task and datasets, while multi-Encoder architectures, such as TNT, are frequently applied to SSC tasks due to their effectiveness in capturing temporal dependencies \cite{135}. Combined with Section \ref{chap:3}. B, GNN-Transformer models are more commonly applied in emotion recognition tasks owing to their ability to model inter-channel relationships \cite{108,111}. In fact, from Section \ref{chap:3}, we also see that generative models, such as GANs and diffusion models, are able to improve data generation and augmentation, mainly addressing challenges in those low-data scenarios. As concluded from Table I and Table A-I, various customized Transformer-based architectures are favorable in emotion recognition, classification, and brain disorders diagnosis tasks. These observed trends highlight some model-task alignments, while without implying rigid generalizations, as EEG decoding tasks often allow for considerable flexibility in the model selection.

Moreover, for the model-dataset matching relationship, we conclude, from the previous tables, that many architectures exhibit similarly strong performance on the same dataset, making it difficult to attribute differences to structural features alone. Indeed, the model performance of the designated dataset is largely driven by the specific optimization strategies and task requirements. One should note that the variability in EEG signal patterns often forces decoding research to focus on single task, resulting in models that are specifically designed and validated for narrow tasks on limited datasets. Most models still lack extensive experimental validation for cross-subject generalization \cite{83,99,108}, and relevant research on task generalization remains limited.

\textit{3) Computational Resources Challenges: }The architecture of transformers and their extensive parameters necessitate signi¬ficant computational resources. To evaluate the computational complexity of a designed Transformer model, the practical metric of the parameter count, reflecting memory needs and hardware compatibility, has been selected. Several representa¬tive data with the explicit record has been reported in supplementary Table D-I. As a result, the Transformer-based decoding networks of EEG has covered different parameter scales of range from K to B. Overall, it is evident that these advanced models exhibit higher parameter counts compared to traditional CNN models (e.g., EEGNet that with mere 1K parameters), implying a more complicated structure. One obvious conclusion is verified that some hybrid models, which belongs to foundation models, improve task generalization but increase computational costs with much larger parameters \cite{64,65}. Apart from this, several CNN-Transformer models balance parameter count by combining shallow CNNs with Transformers \cite{41,87}. Indeed, due to different effecting mechanisms, even models of the same type, such as hybrid models or customized Transformer models, have significantly varying parameters. For instance, some customized models \cite{89,144,173} demand more parameters for higher expressiveness, while another customized model \cite{174} remain compact and effective (i.e., 4.86 K) for focused task-specific applications. In any case, the quest for high-performance models with fewer parameters is always present here.

Beside the parameter size, the computational efficiency of a Transformer model is also coupled with the task requirements and the hardware environment. However, a direct comparison of it across the interested models remains challenging as most studies do not report key metrics, such as the FLOPs, the detailed runtime benchmarks. Thus, we encourage future research to report these additional performance metrics to provide a more comprehensive evaluation. 

\textit{4)  Interpretability Challenges: }There is a need for conti¬nuous exploration of the model’s interpretability. In medical and clinical applications, models with strong interpretability can provide transparency in their decision-making processes, allowing healthcare professionals to verify and understand the outcomes of the models. However, Transformer-based models, despite their excellent performance in handling complex data through numerous parameters and deep structures, often exhibit a high level of abstraction in interpretability, making it difficult to track and explain their decisions \cite{181}. 

\subsection{Future Direction}
The above limitations can be generally categorized as challenges specific to the EEG analysis and those inherent to the Transformer models, which points to two general lines of inquiry for future research. Specifically, from the perspective of Transformer model aspects, the model scaling can be further explored for model parameter compression. Brain Foundation Models \cite{180} trained on extensive EEG datasets and fine-tuned for specific tasks are in their infancy. Therefore, due to the significant demands for computational resources and data scale, small models remain vital in resource-constrained environ¬ments. To realize rapid deployment in specific tasks, it is crucial to develop models with fewer parameters and smaller scales \cite{182}. Customization of the transformer model is also necessary to enhance the training efficiency while reducing computational costs. Moreover, knowledge distillation can be attempted to reduce model size while maintaining performance. Second, the fusion of Transformer with other methods should be investigated to benefit each other. An interesting fact is that there are rare attempts to integrate Transformer with traditional machine learning methods \cite{183}. Although such models may require more experimental validation and optimization, the clear implications of their relevant features deserve further attention to enrich the domain-specific knowledge. Current research on related models is mostly concentrated on the integration with other deep learning models, particularly CNNs, RNNs, and GANs. The combination with emerging models such as GNNs and SNNs, or the development of new Transformer architectures based on these models represents a promising research direction. 

On the other hand, for the EEG signal analysis, future studies could be conducted in the following aspects. First, to avoid the overfitting risk caused by limited data, the dataset expansion is needed. In particular, standardized open-source large-scale datasets, such as the BCI competition dataset, are essential to addressing current Transformer-based model challenges and require special research attention. Large language models and diffusion models have made significant progress in generating text and images. Thus, one can plan to utilize these models to generate EEG data to augment existing datasets. By integrating diverse datasets, Brain Foundation Models can effectively extract universal EEG features, enhancing model robustness and cross-task adaptability. Moreover, tools like Brain Decoder, MOABB, and TorchEEG streamline data preparation and model deployment, addressing the critical need for accessible and widely applied standard datasets. Second, the cross-domain transfer learning should be further emphasized. Applying models pre-trained on text, speech data, and other similar bio-signals to EEG datasets, followed by a fine-tuning, will be a promising method to effectively compensate for the scarcity of EEG training data \cite{30,137}. Third, the multi-modal learning can be further extended. Current studies primarily focuses on the combination of EEG with other bio-signals, such as EOG \cite{131}, \cite{132}, EMG \cite{94}, and electrocardiography (ECG) \cite{160}. In the future, one could explore joint learning of EEG with images, text, or other data to gain a more comprehensive understanding of the characteristics under EEG signal variations. Especially, inspired by GPT and LLaMA, future BFMs may support joint multimodal modeling [184], with advancements in alignment and efficiency paving the way for broader applications in neuroscience.

To accelerate the real-time application of the model, several coupled issues should also be investigated, such as the effect of common EEG noise on designed Transformers, the unified optimization of all hyper-parameters. The real-time processing ability to ongoing EEG is also pretty important and relevant hardware acceleration can be studied. In conclusion, future research should focus on improving the models’ computational efficiency, accuracy, and generalization capabilities to better meet the demands of complex EEG analysis.

\section{Conclusion}
This survey aims to classify and analyze the progress of Transformer-based models in EEG decoding, which adopts a distinguished review methodology that evolves with the model. We classify the related models based on their structural features, specifically including direct backbone models, hybrid models with other DL methods, and customized Transformers. Our analysis focuses on assessing the performance of various Transformer models in processing EEG data, striving to provide the academic community with a comprehensive perspective to foster the development of new technologies. Additionally, it reveals research trends and potential future developments in the field of EEG processing.

\end{document}